\documentclass[letterpaper, 10 pt, conference]{ieeeconf}  %

\IEEEoverridecommandlockouts                              %

\usepackage{relsize}
\usepackage{xspace}
\usepackage{color}
\usepackage{graphics}
\usepackage{graphicx}

\usepackage{amsmath}

\usepackage{amsfonts}
\usepackage{amsmath}
\usepackage{amssymb}

\usepackage[caption=false,font=footnotesize]{subfig}
\usepackage{float}
\usepackage{todonotes}
\usepackage{url}
\usepackage{booktabs}
\usepackage{multirow}
\usepackage{hhline}
\usepackage{hyperref}
\usepackage[capitalise]{cleveref}

\usepackage{todonotes}
\usepackage{tikz}
\usetikzlibrary{arrows.meta}
\usetikzlibrary{calc}
\usetikzlibrary{shadows}
\newcommand{\vspacefiguretikz}{}%
\newcommand{\name}[1]{{\smaller \sf#1}\xspace}

\newcommand{\ROSone}{ROS~1\xspace}
\newcommand{\ROStwo}{ROS~2\xspace}
\newcommand{\rostwotracing}{\texttt{ros2\_tracing}\xspace}

\title{\LARGE \bf
Analyze, Debug, Optimize:
Real-Time Tracing for \\ Perception and Mapping Systems in ROS 2
}

\author{Pierre-Yves Lajoie, Christophe Bédard, Giovanni Beltrame%
\thanks{This work was supported by Vanier Canada Graduate Scholarships, as well as Ericsson, the Natural Sciences and Engineering Research Council of Canada, and Prompt.}%
\thanks{
Department of Computer and Software Engineering,\newline Polytechnique Montréal, Montreal, Canada, \newline
{Corresponding Author: \tt\scriptsize\,pierre-yves.lajoie@polymtl.ca}
}}

\IEEEtriggeratref{8}

\begin{document}

\maketitle
\thispagestyle{empty}
\pagestyle{empty}

\begin{abstract}
    Perception and mapping systems are among the most computationally, memory, and bandwidth intensive software components in robotics. 
    Therefore, analysis, debugging, and optimization are crucial to improve perception systems performance in real-time applications.
    However, standard approaches often depict a partial picture of the actual performance.
    Fortunately, instrumentation and tracing offer a great opportunity for detailed  performance analysis of real-time systems.
    In this paper, we show how our novel open-source tracing tools and techniques for \ROStwo enable us to identify delays, bottlenecks and critical paths inside centralized, or distributed, perception and mapping systems.
\end{abstract} 

\section{Introduction}

Perception and mapping are critical for robot autonomy.
Thus, real-time performance and reliability are essential to perform robust operations and successfully deploy robots outside controlled environments.
Unfortunately, the research community currently lacks efficient diagnosis tools to analyze, debug, and optimize perception and mapping systems.
Those systems are complex and usually distributed over multiple processes such as sensors drivers, data preprocessing, state estimation, visualization, etc.
A global view of the system, which cannot be obtained through traditional debugging, is often necessary to find failure points, deadlocks, and performance bottlenecks.
Researchers and developers are therefore relegated to static code analysis or ad hoc indirect monitoring tools, often with high execution overhead, to troubleshoot and optimize their perception software.
Moreover, collaborative and parallel systems, distributed over multiple computers, add another layer of complexity to the debugging and performance analysis problem.

Software tracing is a practical and effective solution~\cite{gebai2018survey}.
In particular, the LTTng tracer~\cite{desnoyers2006lttng}, which has a mostly negligible overhead on performance~\cite{gebai2018survey}, allows us to create a detailed portrait of the real-time execution of a program. 
It allows developers to detect bottlenecks and measure latencies, among other metrics.
Tracers are an important alternative to classical techniques, such as interactive debugging, when we cannot afford to slow down or pause the execution (e.g., to detect race conditions).
Furthermore, interactive debugging is difficult to deploy on multiprocess systems.

Over the last decade, the \textit{Robot Operating System}, and its successor \ROStwo, have become the most popular software infrastructures for multiprocess management in robotics with their publisher-subscriber messaging scheme~\cite{quigley2009ros,Thomas2014}.
They are the backbone of many robotic applications, and perception systems in particular.

In this paper, we demonstrate how our novel tracing tools and analyses for \ROStwo~\cite{bedard2022ros2tracing,bedard2022message} can benefit the debugging and performance analysis of perception systems.
In particular, we show how important metrics can be extracted from the tracing of a Simultaneous Localization and Mapping (SLAM) system.
We also investigate performance and communication issues in a SLAM system distributed over two different computers.
Overall, the diagnosis and analysis of \ROStwo execution traces could be important tools for large-scale practical deployments of robotic perception such as during the recent DARPA Subterranean Challenge~\cite{darpa_darpa_2020,ginting_chord_2021}.
Code and documentation of the approach are available online: {\small\href{https://github.com/christophebedard/ros2-message-flow-analysis}{github.com/christophebedard/ros2-message-flow-analysis}}.

\section{Background and Related Work}

\begin{figure}[h]
	\centering
	\subfloat[
		\ROStwo trace of a SLAM system.
	]{%
		\fbox{\includegraphics[width=0.97\columnwidth,trim=0mm 0mm 0mm 0mm,clip]{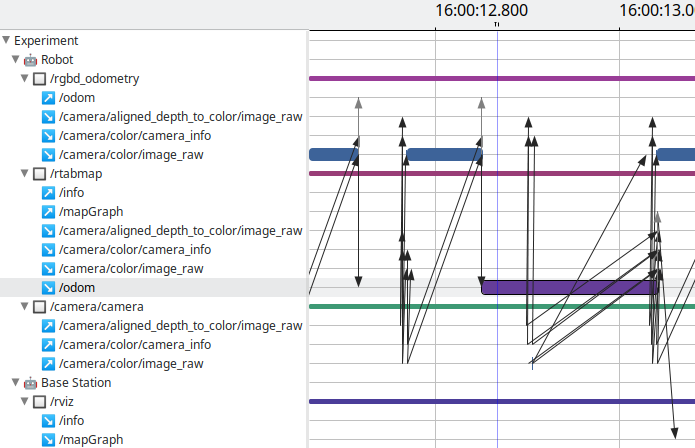}}%
	}
	\hfil
	\subfloat[
		Message flow analysis of an input image.
	]{%
		\fbox{\includegraphics[width=0.97\columnwidth,trim=0mm 0mm 0mm 0mm,clip]{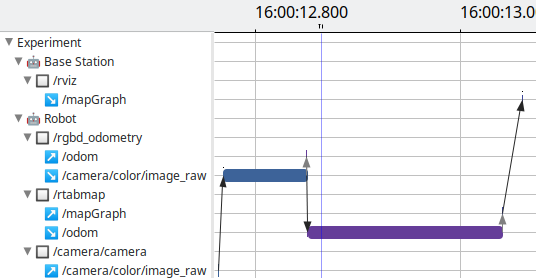}}%
		\label{fig:message-flow-e2e-message-flow}%
	}
	\caption{Visualization of the \ROStwo trace of a SLAM system and the corresponding message flow analysis for an input image. An event timeline is displayed for each \ROStwo topic of each node. Color sections represent callback executions, gray arrows represent links between callbacks and message publication instances, and black arrows show message transfers. This Trace Compass visualization shows the journey of an input image through the SLAM computation pipeline up to the resulting map and pose graph estimate. It enables researchers and engineers to identify delays, bottlenecks, and anomalies.}	
	\label{fig:message-flow-e2e}
\end{figure}

\subsection{Real-Time Perception}

Perception systems, especially those with enhanced capabilities such as semantic understanding, often suffer from performance issues preventing their use in real-time settings.
Thus, there is a trade-off between the system capabilities, from localization accuracy to scene understanding, and the available computing resources~\cite{cadena_past_2016}.
Roboticists also have to consider practical aspects such as sensors data management, and USB bandwidth. 
All those considerations have to be taken into account to achieve real-time performance (e.g., high-speed odometry).
Additionally, in distributed systems, one has to consider data serialization, network delays, synchronization, and coordination which can all be measured with execution traces. 

\subsection{\ROStwo Tracing}

\ROStwo, which is a recent framework already used by many perception systems in robotics, is made of multiple layers.
First, the client libraries, \texttt{rclcpp} and \texttt{rclpy}, offer C++ and Python APIs for developers.
They are based on a common underlying library, \texttt{rcl}, which calls \texttt{rmw}, the middleware interface.
The middleware consists of an implementation of the DDS (Data Distribution Service) standard for dependable, high-performance, interoperable, real-time, and scalable data exchanges in publish-subscribe systems.

The overall performance of \ROStwo is considered superior to its \ROSone predecessor~\cite{maruyama2016exploring,gutierrez2018towards,puck2020distributed}.
To assess this performance, a number of tools have been proposed to evaluate the message transmission latency.
For example, \cite{performancetest} proposes a benchmarking tool to measure the latency between publishers and subscriptions directly, while \cite{irobotros2performance} defines a custom message graph topology.
To enable deeper analysis, several methods have modeled the exchanges of ROS messages from node to node as event chains and pipelines.
They range from online monitoring~\cite{peeck2021online} to  formal scheduling to bound worst-case response times~\cite{casini2019response,tang2020response}.
In a similar line of work, \cite{blass2021automatic} proposed an online automatic latency manager for \ROStwo.

To investigate performance issues in a minimally-invasive way, a tracing tool for \ROSone~\cite{ros1tracetools} has been proposed.
It is based on LTTng~\cite{desnoyers2006lttng}, a tracer with low runtime overhead~\cite{gebai2018survey}.
In previous work, we proposed \rostwotracing~\cite{bedard2022ros2tracing}, a framework with instrumentation and tools for real-time tracing of \ROStwo.
Our subsequent work~\cite{bedard2022message} extends that instrumentation, and proposes an analysis and visualization of the flow of messages across distributed systems.

\subsection{Trace Analysis}

Advanced analysis is required to extract useful pieces of information from tracing data.
In particular, critical path analysis~\cite{yang1988critical}, which searches for the longest execution sequence or pipeline in a distributed system, allows us to compute input-to-output processing time and to identify the bottlenecks.
To achieve this, execution traces from multiple computers need to be synchronized~\cite{poirier2010accurate,jabbarifar2014liana} and then analyzed as a whole.
Inside a ROS system, the execution path takes the form of a message flow in which nodes receiving messages perform some processing, then publish the results to the following nodes, as shown in \cref{fig:message-flow-e2e}.
Then, users can perform a critical path analysis by comparing chains of messages, or pipelines, and identifying the one which takes the longest time between the initial input (e.g., camera images) and the final desired result of the system (e.g., map of the environment). 
To extract the chain of messages, the tracing tool needs to associate input subscriptions with output publishers.
Those causal links can be specified in the application code using instrumentation as in \cite{bedard2022ros2tracing}.
In addition to user-defined annotation, \cite{bedard2022message} offers automatic causal links detection for output messages published directly inside the input subscription callbacks. 
The resulting message flows can be visualized as graphs using the Trace Compass trace analysis framework~\cite{tracecompass}, as shown in \cref{fig:message-flow-e2e}.

\subsection{\ROStwo Tracing for Perception and Mapping}

In this paper, we focus on metrics that are important for the performance of perception systems.
Tracing can help to identify bottlenecks, as well as race conditions, in perception pipelines.
Furthermore, in distributed systems, inter-process and inter-computer communications, including serialization time and network delays, can be measured to analyze the system.
Those communication metrics are especially important since \ROStwo offers several Quality of Service (QoS) settings for messages transmission (e.g., reliability, message queue size, etc.).
Thus, based on the message flow analysis, one could decide which configuration is best suited for a given system or environment.
\section{Experiments}

\begin{figure}[htbp]
    \begin{center}
\begin{tikzpicture}[align=center]
\tikzset{>={Latex[width=1.5mm,length=1.5mm]}}

\begin{scope}[
    every node/.style = {rectangle,draw,node distance=3.1cm,minimum height=0.3cm,inner sep=3pt}
]
    \node[] (a) {Sensor};
    \node[,right of=a] (b) {Odometry};
    \node[,below of=b] (c) {SLAM};
    \node[,left of=c] (d) {Visualization};
    \node[,right of=b] (e) {Motion Planning};
    \node[,below of=e] (f) {Control};
\end{scope}

\begin{scope}[
    baseTxt/.style = {scale=0.80, text centered, font=\sffamily}
]
    \draw[->] (a) -- node [baseTxt, above,sloped] {data} node [baseTxt, below,sloped] {(images, scans, etc.)} (c);
    \draw[->] (a) -- node [baseTxt, above, sloped] {data} node [baseTxt, below, sloped] {} (b);
    \draw[->] (b) -- node [baseTxt, above, sloped] {motion} node [baseTxt, below, sloped] {estimate} (c);
    \draw[->] (c) -- node [baseTxt, above] {map} node [baseTxt, below] {} (d);
    \draw[->,dashed] (c) -- node [baseTxt, above,sloped] {state estimate} node [baseTxt, below,sloped] {map} (e);
    \draw[->,dashed] (c) -- node [baseTxt, above] {state estimate} node [baseTxt, below] {} (f);
    \draw[->,dashed] (e) -- node [baseTxt, above,sloped] {goal} node [baseTxt, below] {} (f);
    
\end{scope}

\end{tikzpicture}
     \end{center}
    \vspacefiguretikz
    \caption{
    Sample representation of the flow of messages in a robotic system. Each box represents a \ROStwo node and arrows correspond to messages transmitted from a publisher to a subscriber. The solid lines represent the perception links studied in our experiments.
    }
    \label{fig:message-flow-graph}
\end{figure}
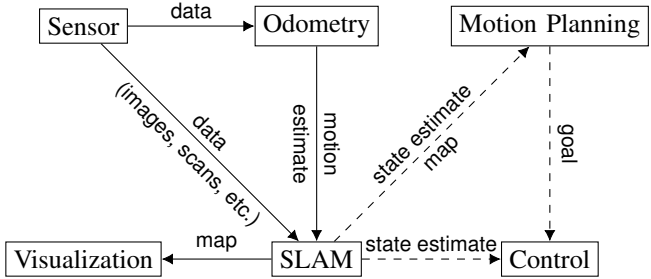

In the following experiments, we show how our tracing tools for \ROStwo and our message flow analysis allow us to extract meaningful performance metrics for perception systems.
In particular, we test our technique on the \ROStwo RTAB-Map SLAM system~\cite{labbe2019}.
We also demonstrate our ability to analyze systems distributed across multiple computers.
The studied SLAM system is composed of four \ROStwo nodes: the sensor driver (Intel Realsense D455 camera), visual odometry, SLAM (loop closure detection, pose graph optimization, etc.), and map visualization.
As shown in \cref{fig:message-flow-graph}, the nodes are linked with message publications and subscriptions, and the raw sensor data is ultimately converted into state and map estimates which can be visualized or used for motion planning and control.
The message flow corresponds to this processing of messages through multiple nodes from the inputs to the desired output.

Causal links between input message subscriptions and output publications are automatically inferred in cases where the processing is done directly in the subscriber callback.
In more complex cases, the links can be defined by the user using a few lines of code.

Our Trace Compass visualization of the message flow displays a graph in which the message publications are linked to the corresponding subscriber callbacks. 
Each \ROStwo topic of each node in the system has its associated timeline in which the execution of callbacks is colored, and the idle time is left blank. 
Transmission of messages from one node to another is represented using arrows from the publication time to the start of the subscription callback. 
Thus, we can move through time, zoom in and out, to inspect the execution trace, measure performance, and look for anomalies.

Our approach also includes the ability to select any message and to compute its associated message flow as shown in \cref{fig:message-flow-e2e}.
Furthermore, the message flow analysis is performed in both directions, i.e., we can select an input to identify the resulting output, or select an output to trace back to the corresponding input.
This enables us to find bottlenecks in the processing pipeline.
Moreover, singling out messages allows us to diagnose fine-grained issues, such as message drops, or an abnormal processing time.
\cref{fig:message-flow-e2e} shows the flow of a camera image message in our SLAM example system.
We can see that the critical path goes through the visual odometry node (\name{rgbd\_odometry}), then the odometry estimate is sent to the SLAM node (\name{rtabmap}) which outputs the resulting map and pose graph to the visualization node (\name{rviz}).
\cref{tab:message-flow-time} presents a breakdown of the end-to-end processing time from the image publication to the resulting map estimate.

\begin{table}[h]
    \centering
    \caption{Time breakdown of the message flow presented in \cref{fig:message-flow-e2e}.}
    \label{tab:message-flow-time}
    \begin{tabular}{lcc}
        \toprule
        \multirow{2}{*}{\textbf{Message Flow Segment}} & \multicolumn{2}{c}{\textbf{Time}} \\
                                                       & \textbf{(ms)} & \textbf{(\%)} \\
        \midrule
        RTAB-Map                           &           191.0  & \phantom{0}64.3\phantom{00}  \\
        Visual Odometry                    & \phantom{0}81.4  & \phantom{0}27.4\phantom{00}  \\
        Network Latency + Message Handling & \phantom{0}24.7  & \phantom{00}8.3\phantom{00}   \\
        Visualization                      & \phantom{00}0.3  & \phantom{00}0.001 \\
        \midrule
        input-to-output total              &           297.4 &            100 \\ %
        \bottomrule
    \end{tabular}
\end{table}

\begin{figure}[h]
	\centering
	\fbox{\includegraphics[width=0.97\columnwidth,trim=0mm 0mm 0mm 0mm,clip]{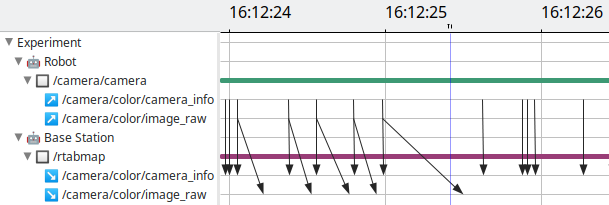}}
	\caption{Visualization of the message publications and subscription callbacks in a distributed SLAM system. In this example, camera images are transferred from one computer to another. The network congestion and bottleneck result in message drops and high varying transmission delays of up to 500 milliseconds. This illustrates that our proposed approach can be used to analyze delays and communication issues in distributed perception systems.}
	\label{fig:network}
\end{figure}

In addition, our approach extends to distributed systems in which multiple traces, synchronized using \name{NTP}\cite{mills1991}, can be analyzed as a whole.
For example, in \cref{fig:message-flow-e2e}, the whole SLAM system is located on onboard the robot and the visualization is done from a base station.
To investigate the system, and showcase distributed systems analysis, we performed another experiment, shown in \cref{fig:network}, in which the odometry is computed onboard the robot, but the SLAM node is executed on the base station. 
While costly in terms of communication, similar workload divisions are common since we can afford to compute loop closure detection and pose graph optimization at a slower rate than odometry.
Thus, we can offload this expensive process to a more powerful computer or server.
Furthermore, offboard computing of SLAM enables collaborative mapping using data from multiple robots simultaneously~\cite{lajoieTowards2022}.
As expected, we can see in \cref{fig:network} that transferring images from the robot to the base station strains the network capacity and introduces delays.
Indeed, the overloaded Wi-Fi network results in varying transmission delays, which include network delays and internal \ROStwo message handling, of up to 500 milliseconds.
We can also observe that only one image message, in the displayed time frame, leads to significant processing in the SLAM node on the base station. 
This indicates the loss of critical mapping information as most images messages are not acted upon.
Hence, our system analysis approach is able to investigate important problems in distributed perception systems such as networks delays and message drops.

\begin{figure}[h]
	\centering
	\fbox{\includegraphics[width=0.97\columnwidth,trim=0mm 0mm 0mm 0mm,clip]{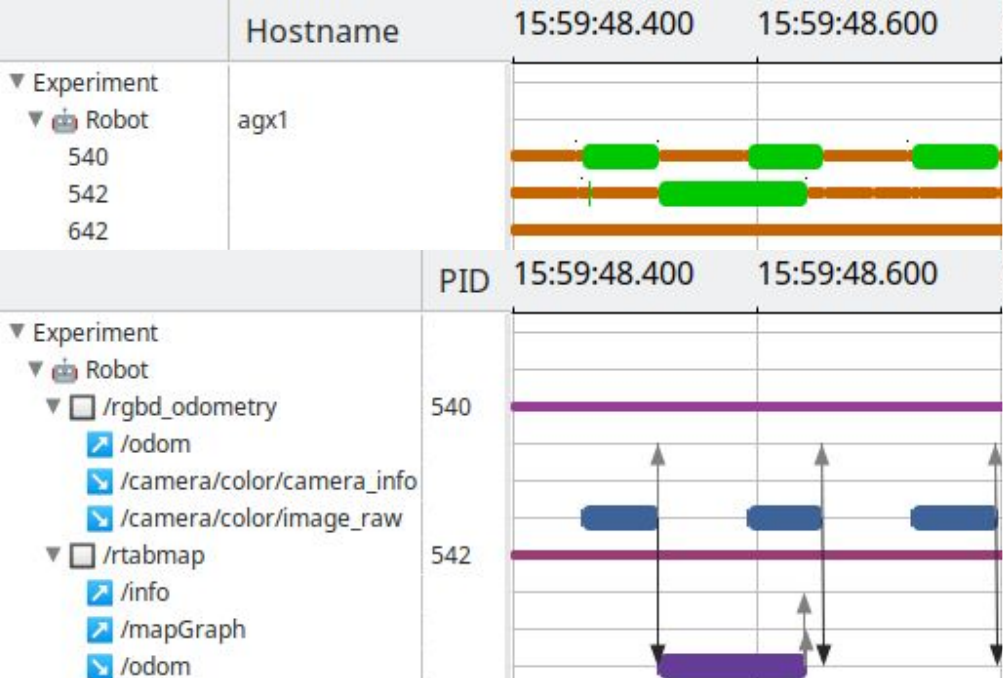}}
	\caption{Visualization of the message publications and subscription callbacks in a distributed SLAM system along with the status of the execution threads (active or idle). The process ID is shown for each \ROStwo node. This combined view allows us to identify computation problems and possible optimizations.}
	\label{fig:executor}
\end{figure}

Besides the message flow, our tracing analysis enables the comparison of message callbacks with the \ROStwo executor threads.
As shown in \cref{fig:executor}, we can see in the top view the status of each thread during the execution. 
Active threads are indicated in green and idle threads in orange.
The process ID indicates which thread is associated with each \ROStwo node (e.g., ID 540 for \name{rgbd\_odometry}).
Thus, we can extract interesting conclusions for the optimization of the systems. 

In light of these results, we believe that our tracing instrumentation of
\ROStwo and our message flow analysis could be important tools for both research
and industry.
\section{Conclusions}

In this paper, we presented how novel real-time tracing tools for \ROStwo enable robotic researchers and engineers to analyze, debug, and optimize SLAM systems.
This is especially important since perception and mapping are often the most computationally, memory, and bandwidth intensive subsystems in robotics. 
Our open-source tracing tool can provide detailed metrics such as end-to-end processing time or data transmission latencies.
The proposed approach is also compliant with distributed systems since we can simultaneously analyze traces acquired on multiple computers. 

Our work on tracing and analysis tools for perception and mapping systems is still in progress.
In the next steps, we plan to develop additional analyses specific for centralized and decentralized SLAM systems including the detection of unused sensor data, and communication bandwidth monitoring.  
We believe that our analysis technique will help to improve the overall robustness of perception systems, and thus facilitate future deployments of resource-constrained robotic systems.

\bibliographystyle{IEEEtran}

\end{document}